\documentclass[preprint,5p,times,twocolumn]{elsarticle}

\usepackage{graphicx}

\usepackage{amssymb}
\usepackage{amsthm}
\usepackage{amsmath}
\usepackage{hyperref}
\usepackage{bm}

\usepackage{multirow}
\usepackage{enumitem}
\usepackage{algorithm}
\usepackage[noend]{algpseudocode}
\usepackage{enumitem}

\begin{document}

\begin{frontmatter}

\title{Distance Metric Learning through Minimization of the Free Energy}

\author[addr1]{Dusan Stosic\corref{cor1}}
\address[addr1]{Centro de Inform\'atica, Universidade Federal de Pernambuco, Av. Luiz Freire s/n, 50670-901, Recife, PE, Brazil}
\cortext[cor1]{Corresponding author.}
\ead{dbstosic@bu.edu}

\author[addr1]{Darko Stosic}
\ead{ddstosic@bu.edu}

\author[addr1]{Teresa B. Ludermir}
\ead{tbl@cin.ufpe.br}

\author[addr2]{Borko Stosic}
\address[addr2]{Departamento de Estat\'{i}stica e Inform\'{a}tica, Universidade Federal Rural de Pernambuco, Rua Dom Manoel de Medeiros s/n, Dois Irm\~{a}os, 52171-900, Recife, PE, Brazil}
\ead{borkostosic@gmail.com}

\begin{abstract}
Distance metric learning has attracted a lot of interest for solving machine learning and pattern recognition problems over the last decades. In this work we present a simple approach based on concepts from statistical physics to learn optimal distance metric for a given problem. We formulate the task as a typical statistical physics problem: distances between patterns represent constituents of a physical system and the objective function corresponds to energy. Then we express the problem as a minimization of the free energy of a complex system, which is equivalent to distance metric learning. Much like for many problems in physics, we propose an approach based on Metropolis Monte Carlo to find the best distance metric. This provides a natural way to learn the distance metric, where the learning process can be intuitively seen as stretching and rotating the metric space until some heuristic is satisfied. Our proposed method can handle a wide variety of constraints including those with spurious local minima. The approach works surprisingly well with stochastic nearest neighbors from neighborhood component analysis (NCA). Experimental results on artificial and real-world data sets reveal a clear superiority over a number of state-of-the-art distance metric learning methods for nearest neighbors classification.
\end{abstract}

\end{frontmatter}


\section{Introduction}\label{secintro}
The success of many machine learning and pattern recognition tasks relies on the choice of an appropriate distance metric, that will be effective in capturing differences between the data. To date, there have been several approaches that attempt to learn distance metrics directly from the input data. Recent works have reported new and more sophisticated methods for learning these distance functions. However, the correct metric choice is problem specific and may dictate the success or failure of the learning algorithm~\cite{Davis}. Distance metric learning consists in choosing the distance metric using information contained in the training data. The resulting distance metrics are widely used to improve performance of metric-based methods, such as nearest neighbors classification~\cite{Cover} or k-means clustering~\cite{Lloyd}. Depending on the problem, appropriate distance metrics can achieve substantial improvements over the standard Euclidean distance metric~\cite{Xiang,Weinberger}. With this in mind distance metric learning plays a crucial role in many pattern recognition tasks, including classification~\cite{Weinberger,Davis}, regression~\cite{Nguyen}, clustering~\cite{Xiang,Xing,Zhang}, feature selection~\cite{Liu,Chang}, and ranking~\cite{McFee}.

Distance metric learning methods can be divided into three categories: supervised, semi-supervised, and unsupervised. Supervised methods for classification aim to ensure that patterns belonging to the same class are close to each other, and those from different classes are farther apart~\cite{Weinberger,Shen}. Semi-supervised methods for clustering use the information to form pairwise similarity or dissimilarity constraints~\cite{Xing,Wang}, and unsupervised methods learn a distance metric that preserves distance relations between most of the patterns for unsupervised dimensionality reduction~\cite{Tenenbaum}. In the supervised context, a popular class of distance metrics is the Mahalanobis distance (see Section~\ref{secbkg}). The distance metric generalizes the standard Euclidean distance by admitting arbitrary linear scalings and rotations. More precisely, learning the distance metric is equivalent to finding an optimal linear transformation that best serves the classification purpose.

While distance metric learning has achieved considerable success in many learning tasks, there remain some challenges to be addressed. For example, the learning process should be flexible enough to support a variety of constraints realized across distinct problem domains and learning paradigms. Distance metric learning approaches based on convex optimizations are formulated under specific constraints, i.e., the matrix must be positive semidefinite, which are unable to generalize well for different tasks. Another challenge is the ability to solve complex constraints, so that the learned distance metric generalizes well to unseen test data. The issue typically arises in distance metric learning methods that take non-convex objective functions with spurious local minima, such as the stochastic neighbors in neighborhood component analysis (NCA)~\cite{Goldberger} which uses gradient descent. This paper focuses on providing a solution to these obstacles.

\begin{figure}[!t]
\begin{center}
\includegraphics[trim = 1in 4.3in 1.4in 0in, clip, width=\columnwidth]{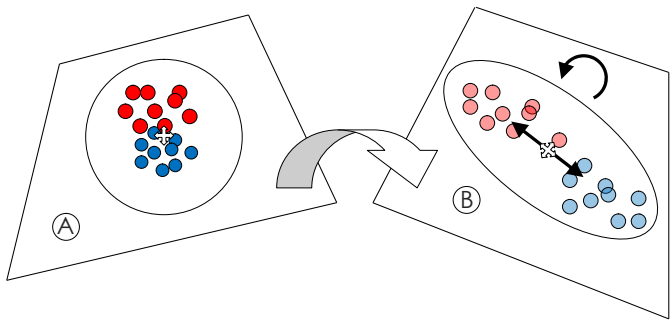}
\caption{Schematic illustration of the intuition behind the learned distance metric. Given a linear transformation $A$, the pattern classes (in red and blue) in the metric space cannot be distinguished (shown in left). Instead a linear transformation $B$ can be learned (through scalings and rotations) such that the distance metric separates the two pattern classes (shown in right).}
\label{figexample}
\end{center}
\end{figure}

For distance metrics learned using complex constraints and non-convex objective functions, an important question entails what is the best distance metric that can be achieved for a given data set? The problem is reminiscent of a typical situation in statistical physics, which allows us to apply its concepts for solving the distance metric learning problem. Rather than performing a gradient-driven iterative search, the basic idea involves applying random transformations on the metric space, i.e., scalings and rotations, in search of a distance metric that minimizes the objective function. This provides a natural way to learn the desired distance metric. The learning process can be intuitively seen as stretching the metric space until some heuristic is satisfied, e.g., similar patterns are closer together and dissimilar patterns are farther apart (see Fig.~\ref{figexample}).

In this paper we introduce a novel approach for learning the distance metric. Our approach differs from other distance metric learning methods in the following ways. We formulate this task as a typical problem in statistical physics. Particularly, we treat the distance metric learning problem as finding the ground state of a complex physical system. The heuristic is identified with the energy of the system, which we minimize by Metropolis Monte Carlo. This allows one to learn the distance metric for non-trivial constraints and non-convex objective functions with spurious local minima. As a case study, we consider stochastic nearest neighbors in NCA, but the same approach can be used for other heuristics. Our numerical experiments reveal that the resulting distance metric leads to significant improvements in accuracy for nearest neighbors classification. Several state-of-the-art distance metric learning methods are used for fair comparison.

The paper is organized as follows. Section~\ref{secbkg} briefly reviews the existing literature on distance metric learning. Section~\ref{secmethod} describes the proposed method. Section~\ref{secexp} demonstrates numerical experiments on nearest neighbors classification. Section~\ref{secdis} discusses important aspects of the proposed distance metric learning method. Section~\ref{secconcl} draws the conclusions.

\section{Related work}\label{secbkg}
Distance metric learning consists in choosing a suitable distance measure from information in the training data. A popular class of distance metrics is the Mahalanobis distance:
\begin{align}
d_M=(u-v)^TM(u-v),
\end{align}
due to its wide use in many application domains and because it provides a flexible way of learning an appropriate distance metric for complex problems~\cite{Kulis,Bellet}. $M$ is a symmetric positive semidefinite matrix that can be factorized into $M=AA^T$, given a linear transformation $A$. The distance $d_M$ between patterns $u$ and $v$ is equivalent to the Euclidean distance after performing the linear transformation:
\begin{align}
d_A(u,v)=(Au-Av)^T(Au-Av).
\end{align}
The distance metric generalizes the standard Euclidean distance by admitting arbitrary linear scalings and rotations (see Fig.~\ref{figexample}), where cross terms (or product between different components) can appear in distance computations through the off-diagonal elements of $M$. The current problem is to learn the linear transformation $u\rightarrow Au$, or its quadratic form $M=A^TA$, that parametrizes the corresponding distance metric. While learning is unconstrained for linear transformations, for the quadratic form we must ensure that $M$ remains positive semidefinite. As a result, most of distance metric learning methods fall into three broad categories: eigenvalue methods based on second-order statistics, convex optimizations over the space of positive semidefinite matrices, and non-convex optimizations.

\begin{figure*}[!t]
\begin{center}
\includegraphics[trim = 0 5.9in 0 0.1in, clip, width=\textwidth]{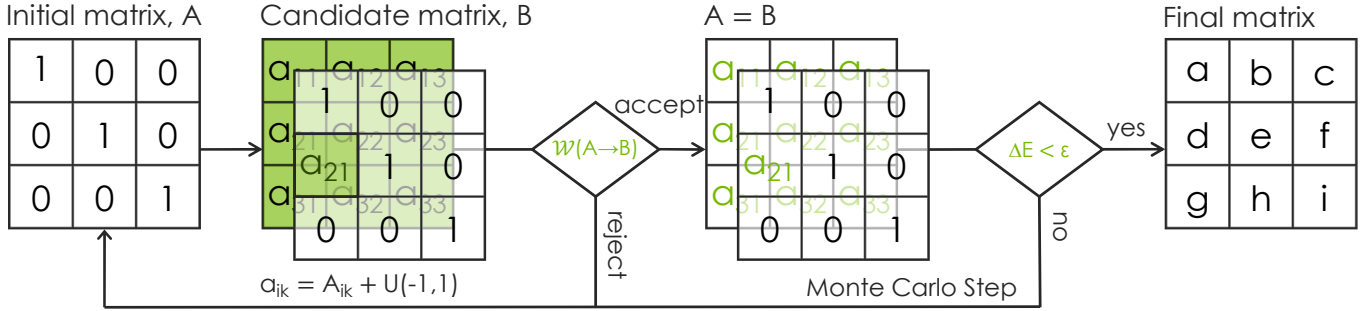}
\caption{Illustration of Metropolis Monte Carlo for distance metric learning. Given the current distance matrix $A$ (e.g., starts with the identity matrix), each Monte Carlo step generates a new candidate matrix $B$ which is accepted or rejected based on the Metropolis choice, $W(A\rightarrow B)$. Candidate matrices are formed by uniformly random deviations $U(-1,1)$ to all elements (background) in $A$. The Monte Carlo step is repeated until some convergence criterion is met, such as when the change in energy, $\Delta E = E_B-E_A$, falls below a given threshold $\epsilon$.}
\label{figmethod}
\end{center}
\end{figure*}

\subsection{Eigenvalue decomposition}
Eigenvalue methods have been widely used to learn the linear transformation since they only need to compute an eigendecomposition. Most of them are based on second-order statistics, such as principal component analysis (PCA)~\cite{Jolliffe}, relevant component analysis (RCA)~\cite{Shental}, discriminant component analysis (DCA)~\cite{Zhao}, and local fisher discriminant analysis (LFDA)~\cite{Sugiyama}. For example, PCA chooses the linear transformation $A$ that maximizes the variance, $\text{max}\,\text{Tr}[A^TCA]$ subject to $A^TA=I$, where $C$ and $I$ are the covariance and identity matrices, respectively. Another important and recent eigenvalue method is DMLMJ~\cite{NguyenMorell}, which learns the distance metric by maximizing the Jeffrey divergence. Such linear transformations can be used either for dimensionality reduction when the matrix is rectangular, or to rotate and re-order the coordinate system for a better representation of the data. They can also be used to accelerate nearest neighbors calculations in large data sets~\cite{Weinberger}. However, eigenvalue methods are often viewed as inducing a distance function rather than learning the underlying metric.

\subsection{Convex optimization}
Distance metric learning can also be formulated as a convex optimization over the cone of positive semidefinite matrices $M$. These approaches are more complicated to devise as they must enforce the constraint that $M$ is positive definite. However, the convexity property makes the problem easier to solve, where there is only one minimum solution that needs to be found. Early works relied on semidefinite programming (SDP) to learn the distance metric under given global constraints~\cite{Boyd}. Ref.~\cite{Weinberger} extended this concept to a large margin setting (LMNN):
\begin{align}\label{eq:lmnn}
E_{LMNN} & = \sum_{j\rightsquigarrow i}d_M(x_i,x_j) \\
&+ \sum_{i=1}^N\sum_{j\rightsquigarrow i}\sum_{\ell=1}^N(1-y_{i\ell})\left[1+d_M(x_i,x_j)-d_M(x_i,x_{\ell})\right]_+ \nonumber
\end{align}
based on a local neighborhood $j\rightsquigarrow i$ that is easier to solve, where $y_{ij}=1$ if the patterns $x_i$ and $x_j$ belong to the same class and $y_{ij}=0$ otherwise, and $\left[\cdot\right]_+=\max(\cdot,0)$. However, standard semidefinite programming~\cite{Boyd} does not scale well when the number of patterns or dimensionality is high. A number of methods have been proposed to reduce the expensive computational cost. Ref.~\cite{Weinberger} suggested an efficient solver based on the projected subgradient descent, while Ref.~\cite{Ying} proposed a method which only requires the largest eigenvalue and corresponding eigenvector to learn the distance metric. Ref.~\cite{Shen} presented a boosting-like method and Ref.~\cite{Shi} formulated distance metric learning as a sparse matrix problem.

\subsection{Non-convex optimization}
Another option is to learn the distance metric through learning a linear transformation. Such distance metric learning methods do not rely on either eigenvalue or convex optimizations. The optimization problem can be efficiently solved by a first-order method, such as gradient descent. However, the problem may be no longer convex and therefore suffers from spurious local minima. As a result, the solution will depend on the initialization point. Neighborhood component analysis (NCA)~\cite{Goldberger} was the first such method, where the linear transformation is learned through direct optimization of the expected leave-one-out classification error:
\begin{align}\label{eq:nca}
E_{NCA}=1-\frac{1}{N}\sum_{i=1}^N\sum_{j=1}^N p_{ij}y_{ij},
\end{align}
from a stochastic neighborhood assignment, $p_{ij}=e^{-d_A(x_{i},x_{j})}/\sum_{k=1,k\neq i}^N e^{-d_A(x_{i},x_{k})}$,
where each pattern $x_i$ inherits the class of a neighbor $x_j$ with probability $p_{ij}$. The key insight to the method is that the distance metric can be found by differentiating Eq.~(\ref{eq:nca}) and using an iterative solver such as gradient descent. NCA was shown to outperform traditional dimensionality reduction and to be capable of improving nearest neighbors classification. This work was later extended under a Bayesian framework~\cite{WangTan} where the posterior probability is encoded in a similarity graph instead of independent pairwise constraints. Ref.~\cite{Davis} formulated distance metric learning as a minimization of entropy by learning a multivariate Gaussian under constraints on the distance function. Other popular methods include large margin component analysis~\cite{Torresani} and multi-task low-rank metric learning~\cite{Yang}.

\section{Proposed Monte Carlo method}\label{secmethod}
The classes of problems in pattern recognition have a close mathematical connection to those in statistical physics~\cite{Richardson}. In relation with pattern recognition, statistical physics has paved the road for much of the research done in academia and industry, ranging from the revival of connectionism~\cite{Hopfield} to modern day deep learning~\cite{Hinton}. Moreover, a parallel can be drawn between statistical physics - the behavior of systems with many degrees of freedom in thermal equilibrium at a finite temperature, and distance metric learning - the search of an appropriate metric by which to measure distances between patterns. As a result, we can employ concepts from statistical physics to learn the distance metric.

The focus of this work is on distance metric learning methods that contain complex constraints and non-convex objective functions, which make them susceptible to getting stuck in spurious local minima rather than the global optimum. For example, the gradient descent used in NCA often converges to local solutions representing suboptimal distance metrics. To solve this issue, we formulate the task as a minimization of the free energy of a statistical system which is equivalent to the distance metric learning problem, under constraints on the distance metric. By analogy with statistical physics, we represent the set of distances between patterns as the constituents of a physical system, where the corresponding energy is given by the constraints or objective function. Since ground states and other low energy states are extremely rare among all possible states of a macroscopic system, we can expect that the methods from statistical physics used to find them are more than capable to search for the global optimum distance metric.

\begin{figure}[!t]
\begin{center}
\includegraphics[trim = 2in 1.5in 2in 0in, clip, width=\columnwidth]{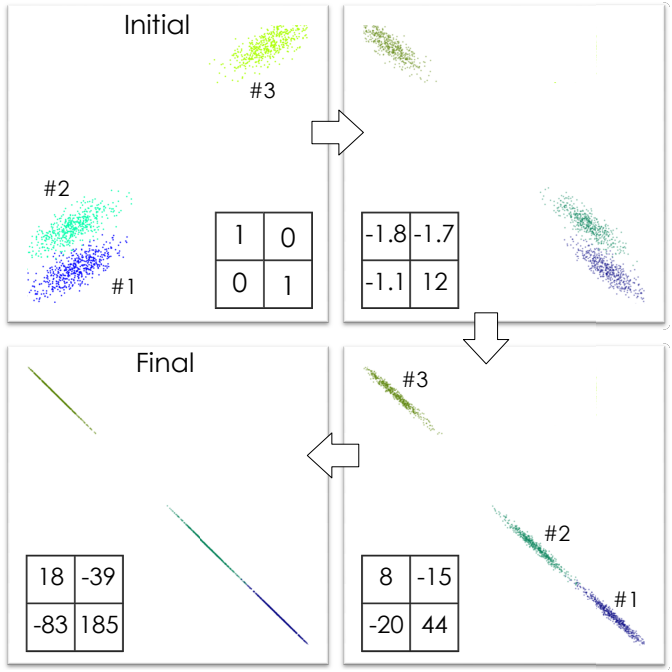}
\caption{Illustration of the distance metric learning process with Metropolis Monte Carlo. Example includes three classes, where the patterns in the learned metric space have opposite orientation and are well separated. Snapshots of the evolution of the metric space starting from the initial Euclidean space and ending with the optimal learned space in clockwise order. The matrices on the bottom denote the learned linear transformations (or scalings and rotations).}
\label{figsimulation}
\end{center}
\end{figure}

\subsection{Problem formulation}
We treat the distance metric learning problem as a search for the ground state of a complex system. The problem is reminiscent of simulations of the thermal motion of particles when the system is in equilibrium, which was solved by random walks in the phase space realized from Metropolis Monte Carlo simulations~\cite{Metropolis}. Here we propose a similar approach to learn the distance metric. 
The basic idea is to devise an iterative search procedure which explores the metric space for the best distance metric given a set of patterns. The nature of the search is important since it must satisfy the detailed balance principle~\cite{Thomsen}:
\begin{align}
\mathcal{P}(A)\mathcal{W}(A\rightarrow B)=\mathcal{P}(B)\mathcal{W}(B\rightarrow A),
\end{align}
which guarantees that a process is both reversible and ergodic. $\mathcal{P}(A)$ represents the probability of the distance metric being in state $A$, with energy $E(A)$ given by Eq.~(\ref{eq:nca}), while $\mathcal{W}(A\rightarrow B)$ is the probability of transition from $A$ to $B$. One possible approach is to make a small random change to the system and calculate the difference in energy, $\Delta E$. This change is then accepted or rejected based on the Boltzmann distribution at a given level of a continuous disorder parameter (equivalent of temperature in statistical mechanics). Repeating this procedure many times allows us to simulate the thermal motion of the distance metric when the system is in equilibrium.

Metropolis Monte Carlo can be used to learn the distance metric (see Fig.~\ref{figmethod} for an illustration) as follows: First an index $i$ is randomly chosen and the matrix element, $A_i$, is displaced randomly to a new candidate value, $B_i$. The difference in energy, $\Delta E=E(B)-E(A)$, is then evaluated and the move is accepted with probability $\mathcal{W}$ given by the Metropolis choice~\cite{Metropolis}:
\begin{align}
\mathcal{W}=\min\left[1,\text{exp}\left(-\frac{\Delta E}{T}\right)\right],
\end{align}
where $T$ is the effective temperature, a disorder parameter in the same units as the cost function~\cite{Kirkpatrick}. The second argument corresponds to the ratio $\mathcal{P}(B)/\mathcal{P}(A)$ of Boltzmann probabilities:
\begin{align}
\mathcal{P}(X)=\frac{e^{-E(X)/T}}{\sum_S e^{-E(S)/T}},
\end{align}
where the normalization factor (or partition function) guarantees that the probabilities sum to one. Only calculations of the energy difference, $\Delta E$, are necessary, while knowledge of the individual state probabilities and 
partition function (which would incur impossible computational costs) is not necessary. Transitions that lead to a decrease in energy
are always accepted, while transitions to higher energies are accepted with a probability that decays exponentially with $\Delta E$ on the scale defined by the choice of $T$.

The procedure is repeated until $N$ trial moves have been attempted, where $N$ is the number of patterns in the system. Each set of $N$ moves comprises a single Monte Carlo step. Repeating this procedure many times allows us to simulate the system in equilibrium, which corresponds to the minimum of the ``free energy":
\begin{align}
\mathcal{F}=-T\log\sum_S e^{-E(S)/T},
\end{align}
and approaches the ground state (or global minimum) as $T\rightarrow 0$. Distance metrics with higher energies (objective function values) are realizable at higher temperatures. For the choice of temperature, we can fix $T=0$ which corresponds to an extremely rapid quenching of the system. The transition probability, $\mathcal{W}$, then acquires a particularly simple form:
\begin{align}
\mathcal{W}=
\begin{cases}
  1, & E(A)\geq E(B) \\
  0, & E(A)<E(B), \\
\end{cases}
\end{align}
yielding a greedy algorithm which only accepts changes to the matrix that reduce the objective function. This makes the approach susceptible to getting stuck in spurious local minima, like gradient descent, but there is no dependence on derivatives. Another possible choice is simulated annealing~\cite{Kirkpatrick} which draws an analogy with annealing in solids, such as crystals. The process consists of first ``melting" the system at high temperatures, and then a very slow cooling occurs where the temperature is lowered in stages (that must proceed long enough for the system to reach equilibrium) until the system ``freezes" and no further changes occur. While the annealing procedure is computationally slower than quenching, convergence to the ground state is guaranteed for sufficiently long times by ergodicity of the dynamical process. The effect of different temperature choices is further discussed in Section~\ref{secdis}.

We refer to the proposed method as Distance Metric Learning through minimization of the Free Energy (DMLFE). A simplified pseudo-code implementation of DMLFE is given in Algorithm~\ref{algo}.

\begin{algorithm}[!t]
\caption{DMLFE}\label{algo}
\begin{algorithmic}[1]
\State \textbf{Initialization:}
\State $\alpha \gets$ annealing parameter
\State $T_0 \gets$ initial temperature
\State $N \gets$ number of patterns
\State $M \gets$ maximum number of iterations
\State $\epsilon\gets$ tolerance for convergence criteria
\State $A = U(0,1)$\Comment{Initialize transformation matrix $A$}
\State \textbf{Monte Carlo procedure:}
\For{$k=1\dots M$}
\State $T\gets \alpha^k\times T_0$\Comment{Temperature annealing}
\State $B\gets A+U(-1,1)$\Comment{Update candidate matrix $B$}
\State $p_{ij}\gets \frac{e^{-d_B(x_{i},x_{j})}}{\sum_{l=1,l\neq i}^N e^{-d_B(x_{i},x_{l})}}$\Comment{Stochastic neighbors}
\State $E_B\gets 1-\frac{1}{N}\sum_{l=1}^N\sum_{j=1}^N p_{lj}y_{lj}$\Comment{Objective func. Eq.~(\ref{eq:nca})}
\State $\mathcal{W}\gets \min\left[1,\text{exp}\left(-\frac{E_A-E_B}{T}\right)\right]$\Comment{Transition probability}
\If{$\mathcal{W}\geq U(0,1)$}
\State $A\gets B$\Comment{Accept update move}
\EndIf
\If{$\Delta E<\epsilon$}\Comment{Convergence criteria}
\State \textbf{break}
\EndIf
\EndFor
\State \textbf{return} $A$\Comment{Output learned distance matrix $A$}
\end{algorithmic}
\end{algorithm}

\subsection{Intuition}
Monte Carlo provides a natural way to learn the distance metric. The method involves applying random transformations on the metric space in search of a distance metric that minimizes the objective function. For example, a candidate matrix $B$ in two dimensions corresponds to the following combination of scaling and rotation transformations of the patterns:
\begin{align}
\begin{bmatrix} x^{\prime} \\ y^{\prime} \end{bmatrix}=
B \begin{bmatrix} x \\ y \end{bmatrix} =
\begin{bmatrix} t_x\cos\theta & -t_y\sin\theta \\ t_x\sin\theta & t_y\cos\theta \end{bmatrix} \begin{bmatrix} x \\ y \end{bmatrix}
\end{align}
where $t_x$ and $t_y$ are the scaling factors, $\theta$ is the angle of rotation. The learning process can then be intuitively seen as stretching and rotating the metric space until some heuristic is satisfied, e.g., similar patterns are closer together and dissimilar patterns are farther apart, as illustrated in Fig.~\ref{figsimulation}.

\section{Experiments}\label{secexp}
In this section, we describe some experiments to evaluate the performance of distance metric learning methods. As a case study, we consider the stochastic nearest neighbors in NCA (see Eq.~(\ref{eq:nca})) for our method, but other non-convex heuristics can be used. The proposed method is compared to the standard Euclidean distance and state-of-the-art distance metric learning methods listed in Table~\ref{tabmethods}, which includes those based on eigenvalue decomposition, convex optimization, and non-convex optimization. Comparisons are also made with Random Forest (RF)~\cite{Breiman}. For our experiments we consider supervised learning for nearest neighbors classification.

\begin{table}[]
\caption{Brief description of methods used in the experiments}
\label{tabmethods}
\centering
\scalebox{0.7}{
\begin{tabular}{ll}
\hline
Method & Description \\
\hline
Euclidean & Euclidean distance metric \\
PCA & Principal component analysis~\cite{Jolliffe} \\
RCA & Relevant component analysis~\cite{Shental} \\
DCA & Discriminant component analysis~\cite{Zhao} \\
LFDA & Local fisher discriminant analysis~\cite{Sugiyama} \\
ITML & Information-theoretic metric learning~\cite{Davis} \\
LMNN & Large margin nearest neighbor classification~\cite{Weinberger} \\
DML-eig & Distance metric learning with eigenvalue optimization~\cite{Ying} \\
SCML & Sparse compositional metric learning~\cite{Shi} \\
NCA & Neighborhood component analysis~\cite{Goldberger} \\
DMLMJ & Distance metric learning through maximization of the Jeoffrey \\
 & divergence~\cite{NguyenMorell} \\
DMLFE & Distance metric learning through minimization of the free energy \\
RF & Random forest [1] \\
\hline
\end{tabular}
}
\end{table}

To make fair comparisons, we consider the following configurations for the experiments. The numerical experiments are empirically tested in the context of nearest neighbors classification, where the number of neighbors is varied between 1 and 40, and the best classification accuracy is chosen. RF is trained with 100 decision trees. We use the source codes of the distance metric learning methods supplied by the authors, and tune their parameters to get the best results. DMLFE uses an annealing procedure which starts with $T=0.1$ and successively reduces the temperature by a multiplicative factor of $\alpha=0.9$. Monte Carlo steps are repeated until convergence or a maximum number of steps is reached.

We compare the distance metric learning methods on 36 data sets from UCI~\cite{Lichmam,AYRNA} and LIBSVM~\cite{LIBSVM} machine learning repositories. The information of these data sets is summarized in Table~\ref{tabdata}. The classification accuracies are obtained by averaging over five runs of two-fold cross-validation. Divisions of the data sets for training and testing are made randomly. The features of the patterns are normalized into the interval $[0, 1]$, which significantly improves performance of many standard distance metric learning methods.

\begin{table*}[]
\caption{Data sets used in the experiments}
\label{tabdata}
\centering
\scalebox{0.75}{
\begin{tabular}{rlrrrrlrrr}
\hline
\# & Data sets & \# of patterns & \# of features & \# of classes & \# & Data sets & \# of patterns & \# of features & \# of classes \\
\hline
1. & Balance			&  625 &   4 &  3 & 19. & Liver				&  345 &   6 &  2 \\
2. & Hepatitis		&  155 &  19 &  2 & 20. & Pima				&  768 &   8 &  2 \\
3. & Ionos				&  351 &  34 &  2 & 21. & Promoters		&  106 & 114 &  2  \\
4. & Iris					&  150 &   4 &  3 & 22. & Soybean			&  683 &  82 & 19 \\
5. & Newthyroid		&  215 &   5 &  3 & 23. & Sonar				&  208 &  60 &  2 \\
6. & Dermatology	&  366 &  34 &  6 & 24. & Vehicle			&  846 &  18 &  4 \\
7. & Breast-w			&  699 &   9 &  2 & 25. & Vote				&  435 &  16 &  2 \\
8. & Breast				&  286 &  15 &  2 & 26. & Vowel				&  990 &  10 & 11 \\
9. & Horse				&  364 &  58 &  3 & 27. & Yeast				& 1484 &   8 & 10 \\
10. & Heart				&  270 &  13 &  2 & 28. & Anneal			&  898 &  59 &  5 \\
11. & Wine				&  178 &  13 &  3 & 29. & German			& 1000 &  61 &  2 \\
12. & Zoo					&  101 &  16 &  7 & 30. & Segment			& 2310 &  19 &  7 \\
13. & Tic-tac-toe	&  958 &   9 &  2 & 31. & Page-blocks & 5473 &  10 &  5 \\
14. & Heart-c			&  302 &  22 &  2 & 32. & Satimage    & 6435  & 36 &  6 \\
15. & HeartY			&  270 &  13 &  2 & 33. & Pendigits   & 10992 & 16 & 10 \\
16. & Card				&  690 &  51 &  2 & 34. & Poker       & 25010 & 10 & 10 \\
17. & Glass				&  214 &   9 &  6 & 35. & Krkopt			& 28056 &  6 & 18 \\
18. & GlassG2			&  163 &   9 &  2 & 36. & Cod-rna			& 59535 &  8 &  2 \\
\hline
\end{tabular}
}
\end{table*}

\begin{table*}[!t]
\caption{Classification error rates for the experiments.}
\label{tabres}
\centering
\scalebox{0.75}{
\begin{tabular}{rrrrrrrrrrrrrr}
\hline
& \multicolumn{2}{c}{\textbf{Classical methods}} & \multicolumn{6}{c}{\textbf{Eigenvalue methods}} & \multicolumn{2}{c}{\textbf{Convex methods}} & \multicolumn{3}{c}{\textbf{Non-convex methods}} \\
\hline
\textbf{\#} &	\textbf{RF}	&	\textbf{Euclidean}	&	\textbf{PCA}	&	\textbf{RCA}	&	\textbf{DCA}	&	\textbf{LFDA}	&	\textbf{DMLeig}	&	\textbf{DMLMJ}	&	\textbf{SCML}	&	\textbf{LMMN}	&	\textbf{ITML}	&	\textbf{NCA}	&	\textbf{DMLFE} \\
\hline
1.	&	15.65	&	10.88	&	10.72	&	9.86	&	11.49	&	13.38	&	9.47	&	8.74	&	8.74	&	9.50	&	9.44	&	6.30	&	\textbf{\underline{5.54}}	\\
2.	&	18.97	&	16.26	&	16.26	&	17.93	&	16.40	&	17.54	&	21.04	&	17.16	&	19.48	&	16.64	&	18.07	&	17.69	&	\textbf{\underline{15.62}}	\\
3.	&	\textbf{\underline{7.18}}	&	14.13	&	14.13	&	15.50	&	17.09	&	11.40	&	14.13	&	11.51	&	14.82	&	12.37	&	12.08	&	13.39	&	12.03	\\
4.	&	4.93	&	3.87	&	3.87	&	2.00	&	4.40	&	9.73	&	4.93	&	\textbf{\underline{2.40}}	&	3.33	&	3.33	&	2.53	&	4.53	&	2.80	\\
5.	&	4.93	&	4.93	&	4.93	&	6.33	&	4.28	&	5.40	&	3.90	&	3.26	&	3.53	&	3.72	&	4.37	&	3.63	&	\textbf{\underline{3.16}}	\\
6.	&	\textbf{\underline{2.24}}	&	3.17	&	3.17	&	3.17	&	2.90	&	23.77	&	9.78	&	2.68	&	3.01	&	3.11	&	3.44	&	3.06	&	3.28	\\
7.	&	3.35	&	3.40	&	3.40	&	4.12	&	\textbf{\underline{2.83}}	&	3.49	&	3.55	&	3.32	&	3.89	&	3.43	&	3.52	&	3.75	&	3.09	\\
8.	&	28.67	&	26.71	&	26.64	&	26.36	&	27.76	&	29.16	&	29.51	&	27.69	&	31.96	&	27.69	&	27.06	&	27.48	&	\textbf{\underline{26.01}}	\\
9.	&	\textbf{\underline{31.81}}	&	36.98	&	36.98	&	37.64	&	35.60	&	38.02	&	36.10	&	33.52	&	38.46	&	34.45	&	37.47	&	34.84	&	34.78	\\
10.	&	19.70	&	16.81	&	16.81	&	17.41	&	17.26	&	19.19	&	18.81	&	17.11	&	20.44	&	16.89	&	16.96	&	17.63	&	\textbf{\underline{16.30}}	\\
11.	&	1.24	&	1.91	&	1.91	&	1.12	&	1.24	&	6.63	&	2.02	&	1.24	&	1.35	&	1.12	&	1.46	&	\textbf{\underline{1.01}}	&	1.91	\\
12.	&	4.76	&	\textbf{\underline{4.16}}	&	4.56	&	23.79	&	4.95	&	59.41	&	19.23	&	4.76	&	5.55	&	6.93	&	5.55	&	5.36	&	4.36	\\
13.	&	2.63	&	0.17	&	\textbf{\underline{0.17}}	&	1.67	&	26.85	&	2.25	&	16.08	&	1.67	&	1.50	&	1.63	&	1.67	&	0.79	&	0.77	\\
14.	&	17.95	&	17.35	&	17.35	&	18.15	&	18.61	&	30.07	&	20.00	&	17.48	&	19.80	&	17.22	&	17.02	&	18.21	&	\textbf{\underline{16.76}} \\
15.	&	17.11	&	16.89	&	16.89	&	16.74	&	17.48	&	18.89	&	18.59	&	16.81	&	19.70	&	16.52	&	18.74	&	17.93	&	\textbf{\underline{15.93}}	\\
16.	&	\textbf{\underline{13.19}}	&	16.12	&	16.12	&	18.29	&	16.00	&	54.06	&	14.99	&	14.26	&	14.58	&	15.33	&	15.36	&	15.71	&	13.30	\\
17.	&	\textbf{\underline{29.63}}	&	35.89	&	35.89	&	35.79	&	37.20	&	41.68	&	39.25	&	37.01	&	36.36	&	37.48	&	38.97	&	36.07	&	34.11	\\
18.	&	\textbf{\underline{16.45}}	&	26.99	&	26.99	&	27.60	&	34.48	&	32.77	&	24.30	&	25.78	&	27.98	&	28.34	&	27.12	&	22.07	&	21.34	\\
19.	&	\textbf{\underline{29.10}}	&	37.62	&	37.62	&	32.52	&	39.89	&	37.45	&	39.48	&	32.98	&	35.19	&	36.53	&	35.77	&	33.63	&	31.42	\\
20.	&	23.91	&	25.55	&	25.55	&	25.29	&	27.63	&	25.76	&	27.66	&	24.43	&	28.02	&	25.63	&	25.34	&	24.56	&	\textbf{\underline{23.07}}	\\
21.	&	\textbf{\underline{16.42}}	&	20.75	&	22.26	&	33.40	&	16.79	&	50.00	&	41.70	&	22.08	&	19.25	&	22.08	&	21.13	&	17.74	&	22.83	\\
22.	&	\textbf{\underline{6.03}}	&	10.25	&	10.28	&	21.58	&	8.76	&	97.07	&	45.83	&	7.12	&	7.26	&	7.06	&	7.35	&	11.39	&	7.96	\\
23.	&	21.25	&	24.52	&	24.52	&	30.87	&	26.83	&	31.73	&	25.10	&	29.42	&	28.94	&	29.33	&	26.92	&	\textbf{\underline{21.15}}	&	21.35	\\
24.	&	19.03	&	25.30	&	25.30	&	16.76	&	35.20	&	31.84	&	28.58	&	20.24	&	\textbf{\underline{18.11}}	&	18.51	&	22.39	&	18.63	&	20.09	\\
25.	&	4.18	&	6.16	&	6.57	&	3.91	&	10.35	&	7.95	&	6.48	&	\textbf{\underline{3.77}}	&	4.69	&	4.60	&	4.97	&	5.06	&	4.23	\\
26.	&	9.82	&	4.69	&	4.69	&	4.32	&	4.32	&	4.97	&	8.51	&	4.10	&	\textbf{\underline{3.92}}	&	4.32	&	4.93	&	5.52	&	4.04	\\
27.	&	\textbf{\underline{40.27}}	&	41.94	&	41.94	&	41.43	&	41.43	&	52.68	&	45.26	&	41.85	&	43.69	&	41.81	&	41.94	&	43.40	&	41.99	\\
28.	&	1.11	&	3.39	&	3.39	&	4.23	&	3.59	&	72.61	&	5.99	&	0.98	&	\textbf{\underline{0.65}}	&	0.85	&	1.20	&	1.94	&	2.38	\\
29.	&	\textbf{\underline{24.64}}	&	27.50	&	27.50	&	28.08	&	26.52	&	29.86	&	29.60	&	26.10	&	28.78	&	26.52	&	27.54	&	27.66	&	26.78	\\
30.	&	\textbf{\underline{2.80}}	&	4.51	&	4.51	&	6.57	&	3.41	&	8.96	&	6.01	&	4.16	&	3.47	&	3.54	&	3.75	&	3.19	&	3.19	\\
31.	&	\textbf{\underline{2.77}}	&	4.24	&	4.24	&	3.76	&	3.98	&	5.12	&	5.02	&	4.47	&	3.88	&	3.76	&	4.13	&	3.91	&	3.44	\\
32.	&	\textbf{\underline{9.36}}	&	9.90	&	9.90	&	17.36	&	10.67	&	10.62	&	11.05	&	9.46	&	11.20	&	9.69	&	9.92	&	9.98	&	9.44	\\
33.	&	1.14	&	0.85	&	0.85	&	0.63	&	1.41	&	1.55	&	1.01	&	0.64	&	0.63	&	0.71	&	0.87	&	0.78	&	\textbf{\underline{0.52}}	\\
34.	&	40.29	&	47.82	&	47.69	&	46.91	&	47.36	&	47.83	&	48.44	&	49.55	&	50.11	&	47.36	&	46.99	&	\textbf{\underline{39.84}}	&	44.16	\\
35.	&	\textbf{\underline{22.22}}	&	35.62	&	35.22	&	36.81	&	36.84	&	36.75	&	62.23	&	34.88	&	33.03	&	35.97	&	37.89	&	30.28	&	30.90	\\
36.	&	4.38	&	7.82	&	7.82	&	4.85	&	22.01	&	5.27	&	18.81	&	22.07	&	20.07	&	6.15	&	4.98	&	5.00	&	\textbf{\underline{4.36}}	\\
\hline
Mean	&	\textbf{\underline{14.42}}	&	16.53	&	16.57	&	17.85	&	18.44	&	27.08	&	16.36	&	16.11	&	15.36	&	16.24	&	21.18	&	17.09	&	14.81	\\
Median	&	14.42	&	15.12	&	15.12	&	17.39	&	16.94	&	24.76	&	13.72	&	13.85	&	14.55	&	15.54	&	18.81	&	16.46	&	\textbf{\underline{12.66}}	\\
Rank	&	4.53	&	6.56	&	6.69	&	7.19	&	8.03	&	11.22	&	7.25	&	5.58	&	6.00	&	5.06	&	10.44	&	7.67	&	\textbf{\underline{3.33}}	\\
\hline
\end{tabular}
}
\end{table*}

\begin{figure}[!t]
\begin{center}
\includegraphics[width=\columnwidth]{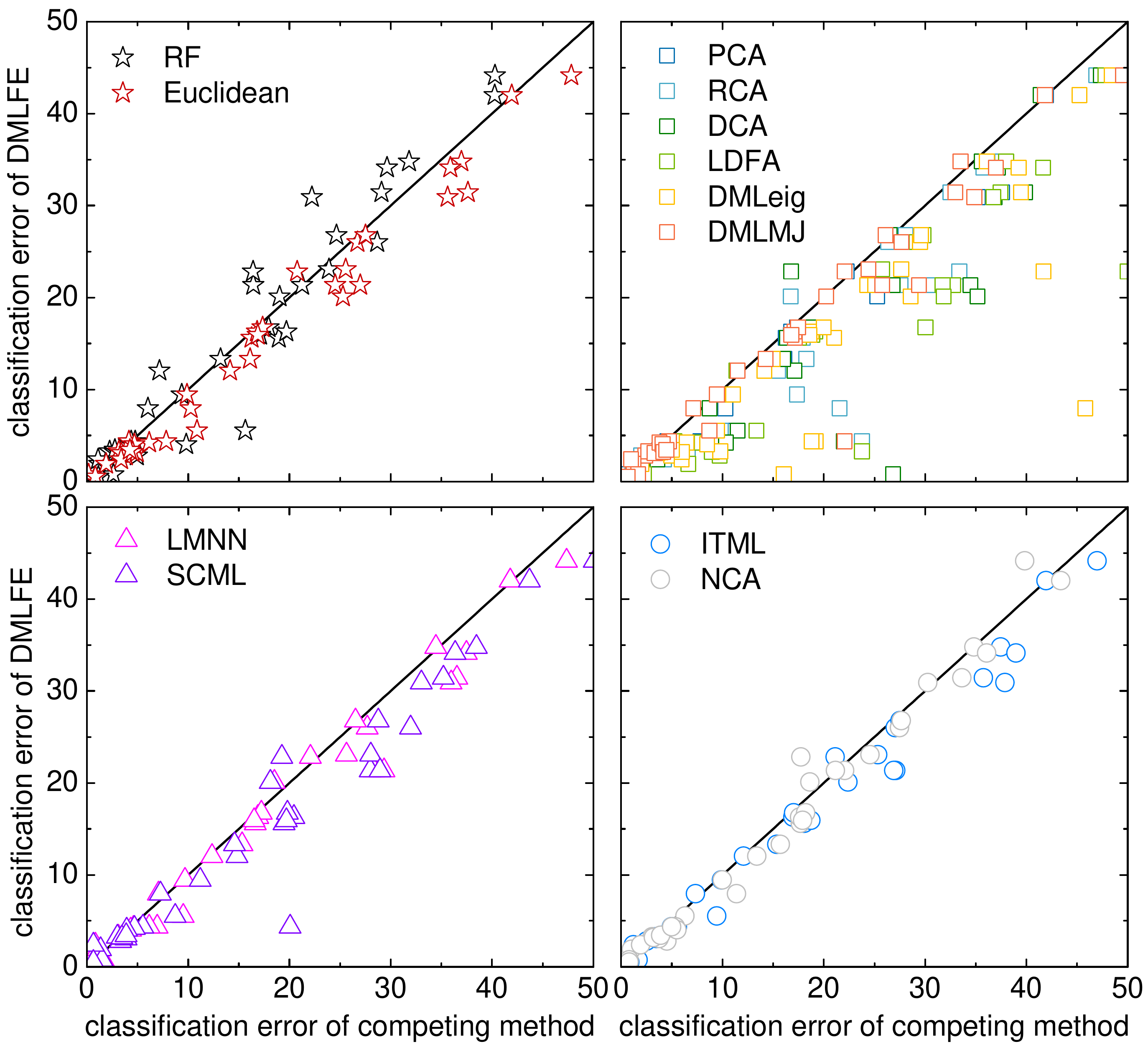}
\caption{Pairwise comparisons of classical, eigenvalue, convex, and non-convex methods for nearest neighbors classification. The coordinates of points represent classification error for each of the two methods compared on single data sets. Points below the diagonal correspond to cases where DMLFE performs better.}
\label{figpairwise}
\end{center}
\end{figure}

Table~\ref{tabres} shows the average classification accuracies obtained by the competing methods. Fig.~\ref{figpairwise} further illustrates side by side comparisons of the results in Table~\ref{tabres}. Since many points lie below the diagonal line, we can conclude that DMLFE presents a clear advantage over other approaches to nearest neighbors classification. Particularly, we find that our methods beats the standard Euclidean distance and is competitive with RF. The approach also seems to widely outperform distance metric learning methods based on eigenvalue decomposition and convex optimization. Lastly, we observe that the Monte Carlo annealing procedure obtains better distance metrics than gradient descent for stochastic neighbors in NCA. The data sets where our method performs poorly can be explained by one of two causes: 1) convergence to a local minimum due to fast annealing, or 2) insufficient Monte Carlo steps. Another plausible reason is that the free energy landscape is not represented by a complex surface with numerous local minima, which occurs in many situations in statistical physics such as spin glasses, but rather by a well-behaved surface with pronounced global minimum, where even gradient-based searches can find the optimal solution.

\begin{figure}[!b]
\begin{center}
\includegraphics[trim = 0 0 0 6.4in, clip, width=\columnwidth]{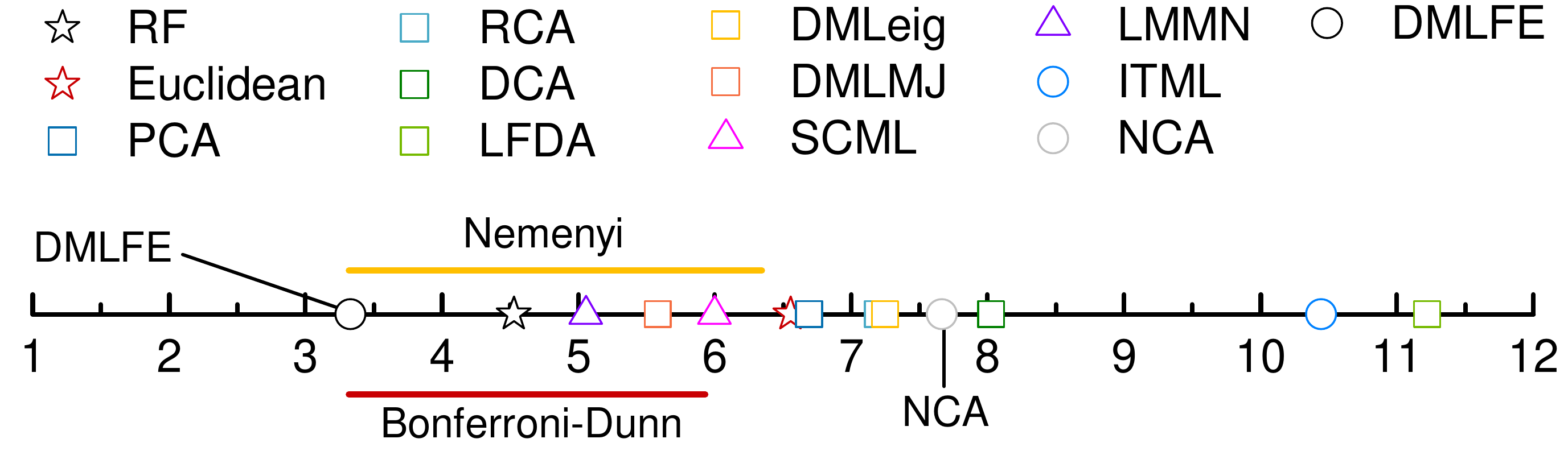}
\caption{Post-hoc tests of the control method (DMLFE) against competing methods. All methods with ranks outside the marked intervals are significantly different from the control.}
\label{figposthoc}
\end{center}
\end{figure}

We rank the methods based on their classification accuracy on each data set, e.g., a rank of one is assigned to the method that has the highest accuracy, a rank of two to the method obtaining the second higher accuracy, and so on. The average ranks and other statistics of the competing methods are listed in the last rows of Table~\ref{tabres}. We find that our approach achieves better results than NCA because the annealing procedure can overcome spurious local minima in which the gradient descent gets stuck. Our approach also significantly outperforms all other distance metric learning methods and has comparable classification performance to random forest (RF), which is often regarded as one of the best classifiers~\cite{Delgado}.

To detect whether there are significant differences among the results, we follow Demsar~\cite{Demsar} for statistical comparisons of classifiers over many data sets. In each case we test the null-hypothesis that the distance metric learning methods obtain the same results on average. We start with the Wilcoxon signed rank~\cite{Wilcoxon} and Friedman~\cite{Friedman} tests. The p-value for the Wilcoxon test is less than the confidence level of $\alpha=0.05$ for all competing methods except RF, so we can reject the null-hypothesis as shown in Table~\ref{tabposthoc}. Similar conclusions can be made about the Friedman test where the p-value is $10^{-23}$. Then we proceed to the Bonferroni-Dunn~\cite{Dunn} and Nemenyi~\cite{Hollander} post-hoc tests to determine which distance metric learning methods perform the same or significantly different from our proposed method. The post-hoc tests can identify significant differences between the control method, which we take to be DMLFE, and other competing methods by computing a critical difference. Two distance metric learning methods are deemed significantly different if their average ranks differ by at least the critical difference~\cite{Sheskin}:
\begin{align}
CD=q_\alpha\sqrt{\frac{n_c(n_c+1)}{6n_t}}
\end{align}
where $q_\alpha$ is the critical value, $n_c=13$ and $n_t=36$ are the number of competing methods and number of data sets, respectively. We consider the Nemenyi and Bonferroni-Dunn tests at a confidence level of $\alpha=0.05$, with critical differences of $3.04$ and $2.64$, respectively. Fig.~\ref{figposthoc} graphically represents the significant differences among the performances of the different distance metric learning methods. Any distance metric learning method outside the marked area is deemed significantly different from the control method, or DMLFE. We find that the post-hoc tests reject the null-hypothesis for all competing methods except DMLMJ, LMNN, and RF as shown in Table.~\ref{tabposthoc}. The statistical results allow us to conclude that our method significantly outperforms competing distance metric learning methods, and its performance is on par with RF.

\begin{table}[]
\caption{Statistical tests for the competing methods with $\alpha=0.05$.}
\label{tabposthoc}
\centering
\scalebox{0.75}{
\begin{tabular}{lrrr}
\hline
\multicolumn{4}{l}{Control method: DMLFE} \\
\hline
\textbf{Method} & \textbf{Wilcoxon} & \textbf{Nemenyi} & \textbf{Bonferonni-Dunn} \\
\hline
RF	&	Accepted	&	Accepted	&	Accepted	\\
Euclidean	&	\textbf{Rejected}	&	\textbf{Rejected}	&	\textbf{Rejected}	\\
PCA	&	\textbf{Rejected}	&	\textbf{Rejected}	&	\textbf{Rejected}	\\
RCA	&	\textbf{Rejected}	&	\textbf{Rejected}	&	\textbf{Rejected}	\\
DCA	&	\textbf{Rejected}	&	\textbf{Rejected}	&	\textbf{Rejected}	\\
LFDA	&	\textbf{Rejected}	&	\textbf{Rejected}	&	\textbf{Rejected}	\\
DMLeig	&	\textbf{Rejected}	&	\textbf{Rejected}	&	\textbf{Rejected}	\\
DMLMJ	&	\textbf{Rejected}	&	Accepted	&	Accepted	\\
SCML	&	\textbf{Rejected}	&	Accepted	&	\textbf{Rejected}	\\
LMMN	&	\textbf{Rejected}	&	Accepted	&	Accepted	\\
ITML	&	\textbf{Rejected}	&	\textbf{Rejected}	&	\textbf{Rejected}	\\
NCA	&	\textbf{Rejected}	&	\textbf{Rejected}	&	\textbf{Rejected}	\\
\hline
\end{tabular}
}
\end{table}

\section{Discussion}\label{secdis}
In this section we outline some important factors which lead to the preferred utilization of DMLFE. We start by considering the choice of temperature, which plays a crucial role in the performance of our proposed distance metric learning method. An annealing procedure will usually guarantee proper convergence to the global minimum, while quenching can get stuck in spurious local minima. However, there are many situations where quenching might provide ``good enough" results and avoid the larger computational costs of annealing. For example, some problems in physics contain many degenerate ground states which are easy to obtain, such as spin glasses, while others are composed of one unique minimum that can be found even with gradient-based searches. One can then imagine that similar situations might occur when learning the distance metric.

\begin{figure*}[]
\begin{center}
\includegraphics[width=\textwidth]{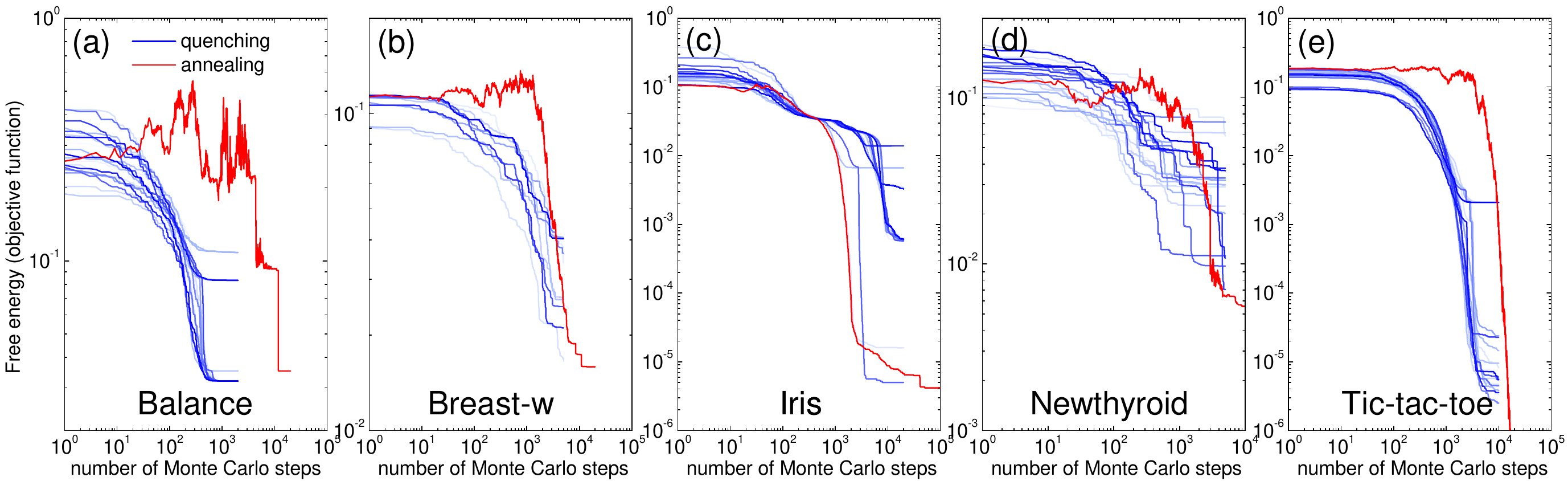}
\caption{Evolution of the free energy for the annealing procedure (red) and for multiple quenching (shades of blue). The quenching procedure is repeated 20 times with different initializations of the distance matrix.}
\label{figtemperature}
\end{center}
\end{figure*}

To analyze the temperature effects we conduct experiments on several of the considered data sets. Fig.~\ref{figtemperature} plots evolution of the energy (or objective function) when using the annealing procedure or multiple quenching with random initialization of the distance matrix. It follows that annealing often results in better solutions (see Figs.~\ref{figtemperature}(b-e)), achieving states that are closer to the ground state in energy, but at the cost of many more Monte Carlo Steps. However, there are situations where multiple quenching can achieve similar or even better results than annealing but with much fewer steps (see Fig.~\ref{figtemperature}(a)). This resembles the case of a frustrated system with nearly degenerate ground states, as mentioned above. While the choice of an annealing procedure with decaying temperature ensures finding the optimal solution, one may still opt to perform several runs with the simpler and faster quenching for different initializations, and then retain only the result with the lowest obtained energy.

\begin{figure*}[]
\begin{center}
\includegraphics[width=\textwidth]{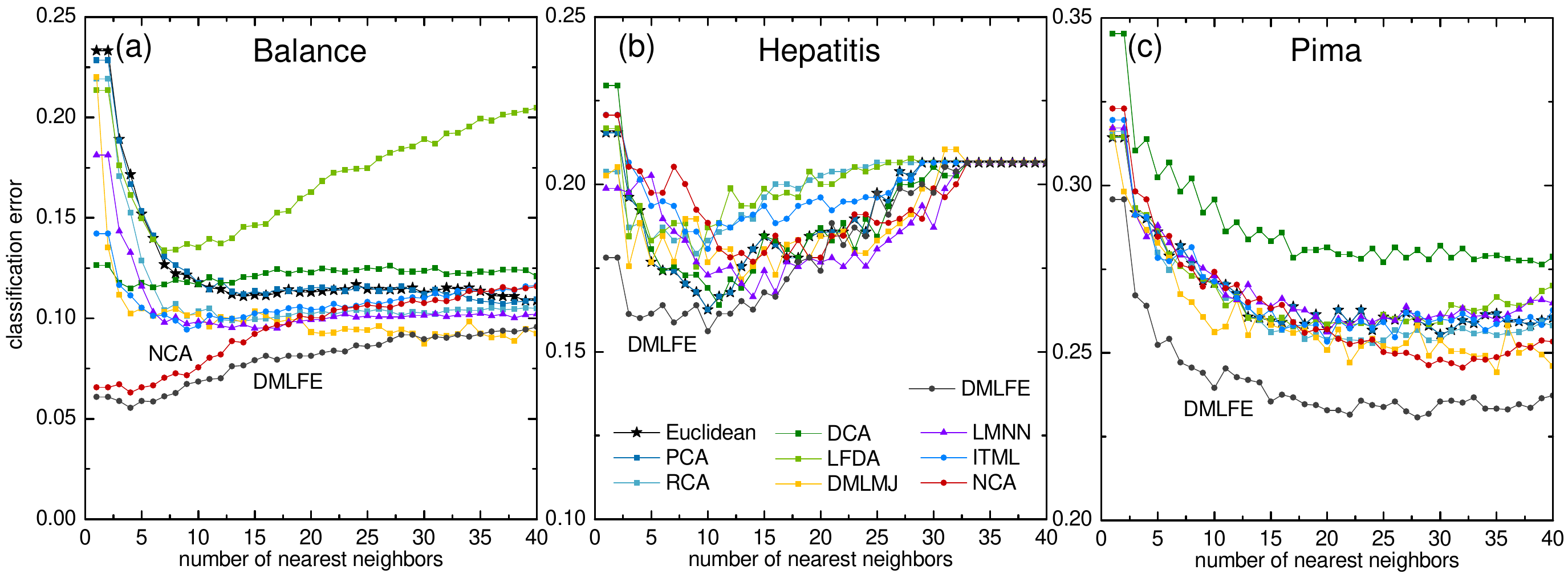}
\caption{Classification error of the competing methods as a function of the number of nearest neighbors.}
\label{figneighbors}
\end{center}
\end{figure*}

We also analyze the performance of competing distance metric learning methods with the number of nearest neighbors. Fig.~\ref{figneighbors} illustrates the classification accuracy as we vary the methods from one to forty nearest neighbors in three of the data sets. We find that our method significantly outperforms NCA for all neighborhoods, which reaffirms our previous conclusion that the Monte Carlo annealing procedure achieves better solutions than gradient descent for stochastic nearest neighbors. In Fig.~\ref{figneighbors}(a), DMLFE achieves better performance than state-of-the-art methods for all neighborhoods, while only for the best performing number of nearest neighbors in Fig.~\ref{figneighbors}(b). Interestingly, in Fig.~\ref{figneighbors}(c) we find that our method achieves better classification accuracy using much fewer nearest neighbors than competing methods based on eigenvalue decomposition and convex optimization. Table~\ref{tabneighbors} reveals similar findings based on statistics over all data sets of the number of nearest neighbors that results in best classification. This suggests that the proposed method with stochastic nearest neighbors can achieve better classification with smaller computational cost than competing distance metric learning methods.

\section{Conclusions}\label{secconcl}
In this paper we developed a novel distance metric learning method. We showed that learning the distance metric can be formulated as a typical problem in statistical physics: distances represent constituents of a physical system and the corresponding energy is given by the leave-one-out classification error in Eq.~(\ref{eq:nca}). Then we demonstrated that this problem is equivalent to minimizing the free energy of a complex physical system. We proposed an approach to find the ground state using Metropolis Monte Carlo simulations. This allows one to learn the distance metric for non-trivial constraints and non-convex objective functions with spurious local minima. As a case study, we considered stochastic neighbors in NCA, but the same approach can be used for other heuristics. The experimental results on synthetic and real-world data sets demonstrate that the proposed method significantly outperforms other state-of-the-art distance metric learning methods for nearest neighbors classification. Future works should be directed towards the construction of  more complex objective functions targeted for specific applications.




\section*{References}
\bibliographystyle{elsarticle-num}

\bibliography{mcdml}

\begin{table*}[]
\caption{Statistics of the number of nearest neighbors that results in best classification for competing methods.}
\label{tabneighbors}
\centering
\scalebox{0.75}{
\begin{tabular}{lrrrrrrrrrrr}
\hline
Measure & Euclidean	&	PCA	&	RCA	&	DCA	&	LFDA	&	DMLMJ	&	LMMN	&	ITML	&	NCA	&	DMLFE	\\
\hline
\textbf{Mean} & 9.77	&	11.97	&	10.93	&	16.63	&	12.73	&	13.4	&	11.6	&	10.5	&	14.3	&	9.3	\\
\textbf{Median}& 6	&	7.5	&	8.5	&	14	&	8.5	&	8	&	8	&	5.5	&	14.15	&	8	\\
\textbf{Rank} & 2.77	&	4.33	&	3.97	&	6.87	&	3.93	&	5.6	&	5	&	3.87	&	5.53	&	3.67	\\
\hline
\end{tabular}
}
\end{table*}

\end{document}